\documentclass[letterpaper]{article}
\usepackage{aaai2026} 
\usepackage{times} 
\usepackage{helvet} 
\usepackage{courier} 
\usepackage[hyphens]{url} 
\usepackage{graphicx}
\usepackage{natbib}
\usepackage{caption}
\usepackage{booktabs}
\usepackage{multirow}
\usepackage{makecell}
\usepackage{flushend}
\nocopyright
\DeclareCaptionStyle{ruled}{labelfont=normalfont,labelsep=colon,strut=off}

\title{LLEXICORP: End-user Explainability of Convolutional Neural Networks}
\author{
    Vojtěch Kůr, Adam Bajger, Adam Kukučka, Marek Hradil, Vít Musil, Tomáš Brázdil
}
\affiliations{
    RationAI,
    Masaryk University,
    Faculty of Informatics,
    Brno, Czech Republic.
}

\begin{document}

\maketitle

\begin{abstract}
Convolutional neural networks (CNNs) underpin many modern computer vision systems.
With applications ranging from common to critical areas, a need to explain and understand the model and its decisions (XAI) emerged.
Prior works suggest that in the top layers of CNNs, the individual channels can be attributed to classifying human-understandable concepts.
Concept relevance propagation (CRP) methods can back‑track predictions to these channels and find images that most activate these channels.
However, current CRP workflows are largely manual: experts must inspect activation images to name the discovered concepts and must synthesize verbose explanations from relevance maps, limiting the accessibility of the explanations and their scalability.

To address these issues, we introduce Large Language model EXplaIns COncept Relevance Propagation (LLEXICORP), a modular pipeline that couples CRP with a multimodal large language model.
Our approach automatically assigns descriptive names to concept prototypes and generates natural‑language explanations that translate quantitative relevance distributions into intuitive narratives.
To ensure faithfulness, we craft prompts that teach the language model the semantics of CRP through examples and enforce a separation between naming and explanation tasks.
The resulting text can be tailored to different audiences, offering low‑level technical descriptions for experts and high‑level summaries for non‑technical stakeholders.

We qualitatively evaluate our method on various images from ImageNet on a VGG16 model.
Our findings suggest that integrating concept-based attribution methods with large language models can significantly lower the barrier to interpreting deep neural networks, paving the way for more transparent AI systems.
\end{abstract}

\section{Introduction}

% TLDR: Introduces CNNs' widespread use in vision tasks and highlights the growing need for explainable AI (XAI) to ensure transparency and trust in high-stakes applications.
Convolutional neural networks (CNNs) remain a cornerstone of modern computer vision~\citep{lecun2015deep,rawat2017deep}, powering applications that range from everyday tasks, such as image classification on mobile devices, to safety‑critical domains like medical imaging~\citep{litjens2017survey} and autonomous driving~\citep{grigorescu2020survey}.
As these models increasingly influence high‑stakes decisions, the demand for interpretable and trustworthy explanations of their predictions has grown~\citep{huang2018trustworthy}.
This need has fueled the development of Explainable AI (XAI), which seeks to make the inner workings and decision rationales of complex models more transparent to both experts and non‑experts~\citep{samek2017explainable}.

% TLDR: Discusses post-hoc XAI methods for CNNs, highlighting popular techniques like Grad-CAM and LIME that explain model predictions without altering the network.
In computer vision, explainability has traditionally been approached by identifying regions of an input image that are most significant for a model’s decision.
Various techniques have been proposed to determine this significance (e.g., Grad-CAM~\citep{selvaraju_grad-cam_2017}, SHAP~\citep{lundberg_unified_2017}, LIME~\citep{ribeiro_why_2016}, or D-RISE~\citep{petsiuk_black-box_2021}).

% TLDR: Explores concept-level interpretability in CNNs, showing how deeper network layers capture human-understandable features (e.g., cat tails or legs) for more intuitive explanations.
Due to limitations of saliency-based explanations~\citep {unreliability_of_saliency_methods_Kindermans2019}, recent research in CNN interpretability emphasizes understanding the higher‑level representations learned by the network.
Studies have shown that neurons or channels in the deeper layers (i.e.,~layers close to the output) of CNNs often correspond to semantically meaningful and human‑understandable concepts~\citep{bau2017network,zhou2018interpreting}.
For instance, a specific feature channel might respond strongly to a cat’s tail, while another might specialize in detecting the contour of its legs.
This concept‑level perspective supports explanations that align more naturally with human reasoning.

% TLDR: Introduces concept attribution methods that link CNN internal features to human-understandable concepts, including Network Dissection, TCAV, and ACE.
A wide range of research in model interpretability focuses on bridging the gap between human‑understandable concepts and the internal representations of neural networks.
Popular methods include Concept Activation Vectors (TCAV)~\citep{kim2018interpretability} and ACE~\citep{ghorbani2019towards}.

% TLDR: Introduces concept attribution methods that link CNN internal features to human-understandable concepts, including Network Dissection, TCAV, and ACE.
A wide range of research in model interpretability focuses on bridging the gap between human‑understandable concepts and the internal representations of neural networks.
Two complementary directions have emerged.
The first is \textit{concept attribution}, which seeks to identify which neurons, channels, or feature maps correspond to semantic concepts.
Popular methods include Concept Activation Vectors (TCAV)~\citep{kim2018interpretability} and ACE~\citep{ghorbani2019towards}.

% TLDR: Explains relevance propagation methods (LRP, DeepLIFT, Integrated Gradients) for tracing prediction contributions, and how combining them with concept attribution enables semantically meaningful explanations like CRP.
The second direction is \textit{relevance propagation}, which identifies which internal components and input regions contribute most to a single prediction.
Methods such as Layer‑wise Relevance Propagation (LRP)~\citep{bach2015lrp, Montavon2019lrp}, DeepLIFT~\citep{shrikumar2017learning}, and Integrated Gradients~\citep{sundararajan2017axiomatic} propagate the prediction score backward through the network to produce fine‑grained attributions.

% TLDR: Introduces Concept Relevance Propagation (CRP) as a method combining concept attribution and relevance propagation, highlighting its strengths and limitations, including reliance on expert manual labeling and difficulty in producing accessible explanations.
In this work, we built upon Concept Relevance Propagation (CRP)~\citep{lapuschkin2023crp}.
In CRP language, a concept is both the semantic concept a channel recognizes and the channel itself.
For a single prediction, CRP finds the most important components for the prediction using LRP. 
Then, the concept is visualized by finding images for which the channel is most relevant during their prediction.

While CRP has demonstrated strong potential for interpreting model behavior, it suffers from two key limitations.
First, the workflow relies on human experts to manually inspect activation images and assign meaningful names to discovered concepts.
Second, translating the resulting relevance maps into coherent explanations requires significant domain expertise, making the method inaccessible to non‑experts and difficult to scale for large‑scale deployment.
Addressing these limitations requires a scalable approach to convert raw concept relevance into coherent, faithful narratives—a challenge well suited to the emerging capabilities of large language models (LLMs).

% TLDR: Introduces LLEXICORP, a framework combining CRP with a multi-modal LLM to automatically generate faithful, human-readable explanations at multiple abstraction levels for both experts and non-experts.
To address these challenges, we propose Large Language model EXplaIns COncept Relevance Propagation (LLEXICORP), a modular framework that integrates CRP with a multi‑modal LLM.
LLEXICORP automatically translates complex relevance maps and concept prototypes into clear, human‑understandable textual explanations while faithfully preserving the underlying CRP reasoning.
The framework supports multiple levels of abstraction, providing technical insights for experts and accessible summaries for non‑specialists, thereby broadening the reach of concept‑based XAI.

Our approach relies on carefully designed prompt engineering.
By supplying the LLM with illustrative examples of CRP outputs, we guide it to produce accurate and coherent narratives.
Moreover, by isolating the concept‑naming and explanation‑generation stages, we maintain faithfulness to the original attribution method while ensuring that each explanation remains interpretable and scalable.

\paragraph{Contribution}
% TLDR: Summary of contributions
In summary, this work (i) identifies the key bottlenecks of CRP for scalable interpretability, (ii) introduces LLEXICORP, a hybrid CRP+LLM framework for automated concept-level explanations, and (iii) demonstrates how prompt-engineered LLMs can make CNN reasoning more accessible to both experts and non-experts.

\begin{figure}[tb]
    \centering
    \includegraphics[width=0.8\linewidth]{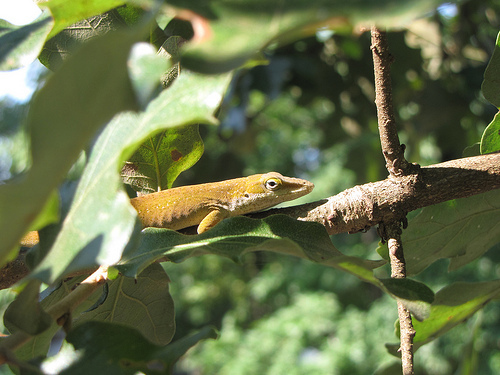}
    \caption{Exaple image of lizard from ImageNet.}
    \label{fig:lizard}
\end{figure}

\begin{figure}[tb]
    \centering
    \includegraphics[width=0.9\linewidth]{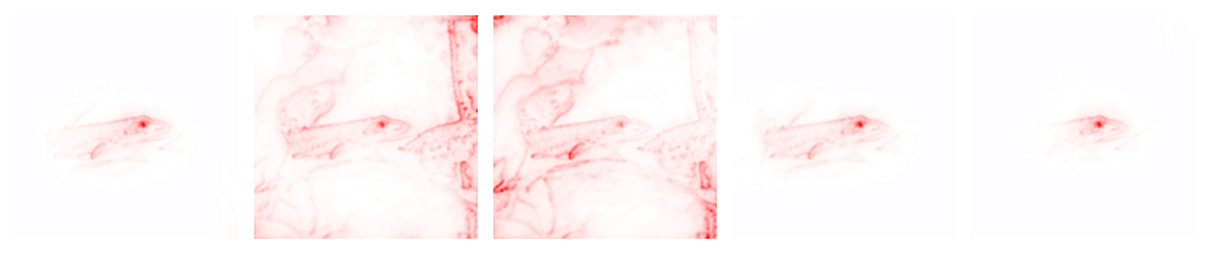}
    \caption{The activation heatmaps of the five most relevant convolutional filters, as found by CRP.}
    \label{fig:lizard_saliency}
\end{figure}

\begin{figure}[tb]
    \centering
    \includegraphics[width=0.9\linewidth]{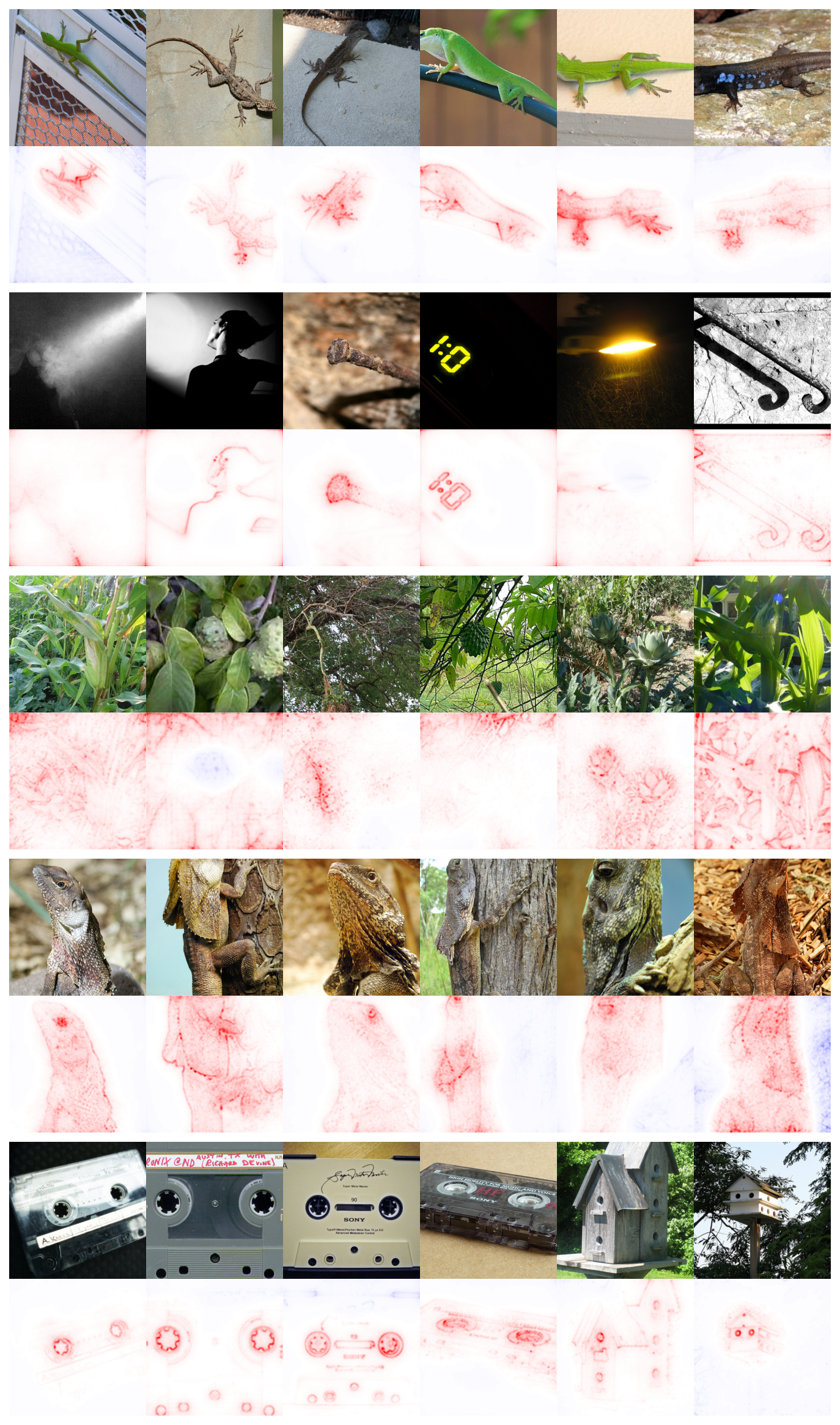}
    \caption{Top 6 images that most activate on each of the 5 most relevant concepts, as found by CRP, together with a saliency map of activation of the said concepts.}
    \label{fig:lizard_representatives}
\end{figure}

\begin{figure}[tb]
\centering
\fbox{
    \begin{minipage}{0.9\linewidth}
The model predicts the image as an ``American chameleon'' with 86.11\% confidence, supported by five key concepts. The most influential concept (40.62\%) is ``lizards,'' directly recognized in the lizard's body, tail, and limbs, which are essential features for identifying the species. The second concept (21.35\%) is ``elongated shapes with distinct edges,' recognized in the lizard's elongated body, head, torso, and tail, contributing structurally to the prediction. The third concept (21.29\%) is ``elongated green leaves or stems,'' directly recognized in the surrounding foliage, which provides contextual association as the natural habitat of the American chameleon. The fourth concept (8.60\%) is the ``rough, scaly texture'' of the lizard's skin, directly recognized on its body and neck, further supporting the identification through its physical features. Lastly, the fifth concept (8.14\%) is ``circular holes or openings,'' recognized in the lizard's eye, which shares visual cues with the concept and contributes to the prediction as a defining physical feature. Together, these concepts explain the model's confident classification of the image as an American chameleon.

    \end{minipage}
}
\caption{LLEXICORP generated summary for the image of the lizard and given the outputs of CRP.}
\label{fig:model-result}
\end{figure}

\paragraph{Example} As an example, we use an image of a lizard from ImageNet \citep{deng2009imagenet}, which is used in the notebook provided by \citep{achtibat_attribution_2023}.
The image is shown in Figure~\ref{fig:lizard}.
The CRP finds the top 5 most relevant concepts for the prediction of the image as an  American chameleon.
The maps that show where these concepts activate are shown in Figure~\ref{fig:lizard_saliency}.
For each of the concepts, the top 6 images that activate the most for that concept are shown in Figure~\ref{fig:lizard_representatives}, together with saliency maps of the concepts' activation.
Finally, the summary generated by LLEXICORP is provided in Figure~\ref{fig:model-result}.
As we can see, LLEXICORP successfully names what the individual concepts are, what their role is in the image, and explains how that relates to the prediction.

\section{Related Work}

% TLDR: Highlights the growing role of LLMs as evaluators and interpreters in ML and XAI pipelines, emphasizing their ability to translate complex model outputs into natural-language explanations that enhance accessibility and interpretability.
Large language models (LLMs) have increasingly been integrated into diverse machine learning pipelines, where they serve as evaluators, assistants, and interpreters.
They have demonstrated strong alignment with human judgments on open‑ended tasks, enabling scalable evaluation without costly manual labeling~\citep{zheng_judging_2023}.
Models such as LIMA further showed that even lightly fine‑tuned LLMs can generate high‑quality, human‑preferred outputs across various domains~\citep{zhou_lima_2023}.
Building on these capabilities, LLMs are now being explored in explainable AI (XAI) pipelines, where they translate complex model outputs—such as saliency maps or concept activations—into coherent natural‑language narratives.
This integration lowers the barrier for non‑experts while enhancing the interpretability and accessibility of model explanations~\citep{bilal_llms_2025,martens_tell_2025,ziems_explaining_2023}.

Early research explored the use of large language models (LLMs) to translate machine learning outputs into human‑readable narratives. 
\citet{susnjak_beyond_2024} demonstrated how ChatGPT can convert counterfactual explanations from a learning analytics model into structured natural‑language recommendations, though they did not evaluate user comprehension. 
Similarly, \citet{ali_huntgpt_2023} applied LLMs in network anomaly detection, generating conversational narratives from SHAP and LIME outputs to enhance accessibility. 
Other studies have focused on simpler models: \citet{lorig_exploring_2024} and \citet{ziems_explaining_2023} leveraged ChatGPT to explain decision tree predictions in small‑scale or domain‑specific tasks, highlighting the promise of LLMs as explanatory intermediaries but offering limited assessment of fidelity and scalability. 

Beyond these isolated applications, several frameworks have emerged to systematically integrate LLMs into XAI workflows. 
\textit{XAIstories} produces natural‑language narratives such as SHAPstories and CFstories to improve user comprehension, while \textit{x‑[plAIn]} adapts explanations to audience expertise for greater accessibility \citep{martens_tell_2025, mavrepis1_xai_2024}. 
Evaluation frameworks such as \textit{VirtualXAI} further combine quantitative benchmarks with qualitative user assessments by simulating diverse user personas \citep{hosseini_role_2025}. 
Recent surveys provide a comprehensive view of LLM‑driven XAI, noting both the potential for enhanced interpretability and the risks of hallucinations, bias, and oversimplification in generated narratives \citep{zhuang_large_2025, kostic_llms_2024, atkinson_llm-based_2025, thelwall_evaluating_2025, dreyer_revealing_2022}. 

AI research increasingly pivots towards the autonomous, goal-driven realm of agentic AI \cite{hosseini_role_2025}.
LLEXICORP leverages LLMS to translate complex (CRP) outputs into human-understandable narratives and follows the agentic design pattern by combining two agents to form an explainable, user-friendly system.

\section{Background}

We consider a single convolutional layer of a convolutional network.
Let $C$ denote the set of all convolution filters of the layer.
Our method is designed to interpret the outputs of explanation techniques, such as CRP, that identify concepts using convolutional filters and assess their relevance to the network's particular input-output behavior.

For our method, we assume two external methods: one, which we call \emph{concept identifier}, and the other, \emph{concept visualizer}.

For a fixed image $I$ and output class $y$, a \emph{concept identifier} computes a set of channel $\{c_1, \ldots, c_n\} \subseteq C$ which represent the most relevant channels for identification of image $I$ as class $y$ and their relevance $p_1, \ldots, p_n$ which state how relevant the concepts are for predicting $I$ is $y$.
Furthermore, for each $i \in \{1, \ldots, n\}$ the method provides an overlay mask (or heatmap or saliency map) $h_i$ over $I$ indicating the position of the concept $c_i$ in $I$.

For a fixed concept $c_i$ (a convolutional filter), the concept visualizer provides a set of images $r_{i}^{1},\ldots,r_{i}^{k}$ that best represent the concept.
Concept Relevance propagation is a method that offers both of these things.
\section{The Method}

Here, we describe the inner workings of our method.
On input, we accept an image $I$ and an output class $y$, the model's confidence in that class $p$.
We use the concept identifier to identify concepts $c_1, \ldots,c_n$, saliency maps $h_1, \ldots, h_n$ and relevance numbers $p_1, \ldots, p_n$.

We pass each concept $c_i$ to the concept visualizer to retrieve the set of representative images $r_{i}^{1},\ldots,r_{i}^{k}$.
LLEXICORP then instructs the LLM to describe the common visual pattern of the concepts' representatives to get a $l_i$--a textual label of the common pattern.

Then, for each $i$, LLEXICORP sends the original image $I$, class $y$ and confidence $p$, heatmap $h_i$, relevance $p_i$ and label $l_i$ to the LLM to explain (a) how that visual pattern is recognized in the image, given the description of the pattern $l_i$ and the position in the image $h_i$, (b) how that relates to the prediction of $I$ as $y$, using also the relevance of the concept and confidence of the prediction.
This way, we achieve a description of the concepts contextualized with respect to the original image and the prediction.
Let us call the output of this prompt $d_i$.

Finally, the LLM summarizes the contextualized concepts $d_i, \ldots, d_n$ into a concise summary.

In implementation, we used OpenAI's gpt-4o-2024-11-20~\citep{openai_gpt4o_2024_11_20} and Concept Relevance Propagation as provided in the default implementation found \citep{anders2021software}.
As per the CNN, we use a pretrained VGG16 model in Pytorch~\citep{pytorch_vgg16_bn}.

We will now describe each LLM instruction (prompt) separately.
The full prompts are in the appendix, along with an ablation study of the effects of individual parts of the instruction messages.

\subsection{Concept Labeling}

The first prompt is for the concept visualizer to name the common visual pattern.
The instruction prompt first states that it is an AI assistant explaining CRP.
We specifically explain that the pattern must be recognizable by a convolutional filter.
This is important because simply stating the common feature leads the LLM to also consider inner properties.

Because we use CRP, where each $r_i^j$ is a pair of images—an image from the validation set and an overlay heatmap indicating its position—we also include an explanation of the heatmap.

\begin{figure}[tb]
    \centering
    \includegraphics[width=0.9\linewidth]{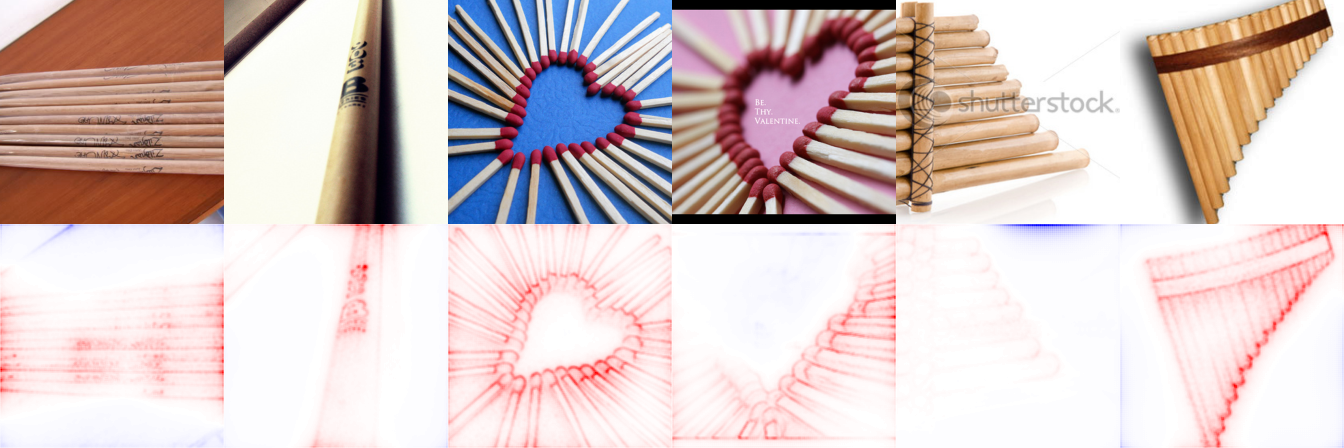}
    \caption{An example of a set of representatives of a concept with their saliency maps provided by the concept visualizer.
    The result of LLM's annotation is: \emph{The concept appears to be ``wooden cylindrical objects arranged in parallel or patterns.'' This includes items like wooden sticks, matchsticks, or pipes, often aligned or forming shapes.}
}
\label{fig:sticks}
\end{figure}

For example, the last row in Figure~\ref{fig:lizard_representatives} is described as:
\emph{The concept appears to be ``circular holes or openings with a distinct rim.'' These are seen in the cassette reels of the tapes and the circular openings in the birdhouses. The pattern is characterized by round shapes with a defined edge or border.}
Another example is shown in Figure~\ref{fig:sticks}.

\subsection{Contextualization}

In the second prompt, we aim to identify how that concept is present in the original image.

Here, we give the LLM two main rules: one set of rules for concept-pattern relation and the second for concept-prediction relation.

This first part is about determining the most fitting relation between the concept description and the highlighted pattern, and what the recognized concept:
\begin{enumerate}
    \item  Direct recognition: the concept and the pattern are identical, even if the pattern is only a part of the image (concept is snow, pattern is snow), the recognized concept is the concept itself.
    \item Feature recognition: the concept and the pattern share visual cues (pattern is a leg of a dog, concept is a leg, or pattern is a tail of a husky, concept is a tail of wolves), the recognized concept is the pattern in the image.
    \item Co-occurrence recognition: The concept is not present in the highlighted region, but the highlighted pattern or the whole object often appears with the concept in real life (for example, beehives and bees, steak and potatoes). The recognized concept is the concept itself.
    \item  Misidentification: The concept is not present in any form. This means that the concept is not relevant to the prediction.
\end{enumerate}
We also provide three decision rules:
\begin{itemize}
    \item  If the highlighted region matches the concept description, choose direct or feature recognition, even if the concept is part of the background or a part of the predicted object (given an image of a camel on sand and sand concept, highlighted part is the sand, then this is a direct recognition of the sand, not co-occurrence).
    \item Even if the concept is not present in the image, check for co-occurrence, there might a learned association between the concept and the pattern.
    \item Use co-occurrence over feature recognition if the shared visual cues are weak (given an image of a camel on grass and concept sand, camels being on sand is strong co-occurrence, the recognized concept is the sand).
\end{itemize}

The second set of rules is for the concept-prediction relation.
If the concept is recognized in the image, the LLM should determine how the recognized concept relates to the prediction.
\begin{itemize}
    \item Exact classification: The recognized concept itself is the predicted class.
    \item  Compositional association: The recognized concept is a physical part, structural feature, or natural arrangement of the predicted object (such as an eye, leg, wing, or horseshoe as a C-shaped object). It does not need to be a unique or defining feature of the class; any part of the object.
    \item Contextual association: The recognized concept is not part of the object but commonly co-occurs with it (such as snow for a husky, road for a car). The association does not need to be a defining feature of the predicted class. 
    \item Misassociation: The recognized concept is unrelated to the prediction, possibly an issue in the model. 
\end{itemize}

We have found that splitting the prompt into two tasks was conceptually easiest for the LLM.
This is likely because this way, the prompt functions more like a decision algorithm.

For example, continuing with the last row in Figure~\ref{fig:lizard_representatives} which has the saliency map in Figure~\ref{fig:lizard_saliency} as the last one, the contextualization states this:
\emph{The concept is described as ``circular holes or openings with a distinct rim,'' characterized by round shapes with a defined edge or border. The saliency map highlights the eye of the lizard in the image. The concept and the highlighted region share visual cues, as the lizard's eye has a circular shape with a defined edge, making this a feature recognition of the lizard's eye. The recognized concept, the lizard's eye, relates to the prediction of ``American chameleon'' through compositional association, as the eye is a physical part of the lizard and contributes to its identification.}

For the third row in the picture with the foliage, we get this response:
\emph{The concept is described as ``elongated green leaves or stems,'' which refers to narrow, pointed plant structures. The highlighted region in the saliency map corresponds to the leaves and stems surrounding the American chameleon in the image. This is a direct recognition of the concept, as the highlighted pattern matches the description of elongated green leaves or stems.
The recognized concept relates to the prediction through contextual association. While the leaves and stems are not part of the American chameleon itself, they provide a natural environment where this species is commonly found. The model likely associates the presence of such foliage with the habitat of the American chameleon, contributing to its prediction.}

\begin{figure}[tb]
    \centering
    \includegraphics[width=0.9\linewidth]{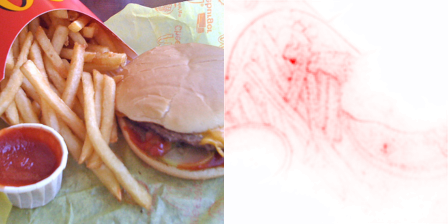}
    \caption{An example of an image and concept activation of the third most relevant concept, as found by the concept identifier.}
    \label{fig:fries}
\end{figure}

For the images of sticks in Figure~\ref{fig:sticks}, this was the third most relevant concept in a picture of a cheeseburger, with concept activation shown in Figure~\ref{fig:fries}.
Here, we get a result: \emph{The concept is described as ``wooden cylindrical objects arranged in parallel or patterns,'' such as wooden sticks or matchsticks. The saliency map highlights the region of the image containing the French fries. The concept and the highlighted region share visual similarities, as the elongated, cylindrical shape and parallel arrangement of the fries resemble the described concept. This is a case of feature recognition, where the recognized concept is the French fries in the image. The recognized concept (French fries) relates to the prediction of ``cheeseburger'' through contextual association, as French fries are commonly served alongside cheeseburgers, reinforcing the model's confidence in its prediction.}

\subsection{Summarizing}

The last part involves simply summarizing the conceptualized labels into a concise summary.
We emphasize that the image or concept representatives are not sent again here, only the text outputs.

In the prompt, we instruct the LLM to summarize the text without adding new information and to use non-technical language.

\section{Evaluation Questionnaire}

To assess the method's effectiveness in bringing CRP outputs closer to regular users, a qualitative study is conducted.
With a fixed seed of 0, we sampled 8 images from the ImageNet validation set \citep{deng2009imagenet} and generated explanations for each.
We then asked participants to rate the quality and helpfulness of the generated texts relative to the images, predictions, identified concepts, their representative images, and the CRP saliency maps.

We use images from ImageNet because it is a natural continuation of work on CRP.

We received 21 answers from voluntary participants. 

The performance evaluation of LLEXICORP is conducted through a comprehensive online questionnaire that gathers user feedback on the generated textual explanations. The study's primary objective is to assess how effectively LLEXICORP converts the outputs of the Concept Relevance Propagation (CRP) XAI technique into concise, human-readable textual explanations using an LLM.

\subsection{Questionnaire design}

The questionnaire is structured to guide evaluators through a systematic assessment of LLEXICORP's output for multiple images. Each image receives a dedicated evaluation section, clearly separated by a page break to ensure clear delineation and prevent cognitive overload between evaluations.

Each image evaluation section begins with an ``Evaluation Instructions'' overview, detailing the three main steps of the assessment process:
\begin{enumerate}
    \item \textbf{Image and Model Prediction}: Participants first review the original image and the classification model's prediction.
    \item \textbf{Individual Concept Evaluation}: Participants then evaluate individual ``concepts,'' which are visual patterns the neural network has learned to recognize. For each concept, the form displays a saliency map highlighting relevant parts of the original image, representative images (each with its own saliency map), and a textual description of the concept generated by LLEXICORP. Participants rate four aspects: whether the description accurately captures the visual pattern in the representative images, whether the description identifies the highlighted areas in the original image, whether the explanation of the concept's presence in the image is reasonable, and whether it is useful for understanding the prediction.
    \item \textbf{Final Summary Evaluation}: Finally, participants review the image and prediction again, along with a comprehensive summary that combines all individual concept explanations. They evaluate whether the summary contains only information from the individual concept descriptions, whether it is helpful for understanding the prediction, and whether it is more helpful than the individual concept descriptions.
\end{enumerate}

Following these instructions, the evaluation for each image proceeds through the three steps:

\subsubsection{Step 1: Image and Model Prediction}
This initial section presents the core visual and textual input for evaluation:
\begin{itemize}
    \item \textbf{Original Image}: The image that was provided as input to the classification model.
    \item \textbf{Model Prediction}: The classification label predicted by the model for the original image.
\end{itemize}

\subsubsection{Step 2: Individual Concept Evaluations}
This section is generated based on the five most relevant concepts identified by CRP for the specific image. Each concept's evaluation is visually separated by a header (e.g., ``--- Concept 1 of X ---''). For each concept, the following visual and textual information is presented:
\begin{itemize}
    \item \textbf{Saliency Map for Concept [Concept Number]}: A heatmap overlaid on the original image, indicating the regions most relevant to this specific concept.
    \item \textbf{Representative Images for Concept [Concept Number]}: A collection of images that strongly activate this concept, each accompanied by its corresponding saliency map, indicating what parts of the representative image signify the concept.
    \item \textbf{Concept Description}: The textual narrative generated by LLEXICORP (via ChatGPT), explaining what the concept is and how it is represented in the representative images.
\end{itemize}
After reviewing these materials for each concept, participants respond to four multiple-choice questions using a uniform rating scale of `Agree', `Not sure', or `Disagree':
\begin{enumerate}
    \item ``The concept describes a common visual pattern highlighted in the representative images.''
    \item ``The description identifies the highlighted areas in the original image.''
    \item ``The explanation of the concept's presence in the image is reasonable.''
    \item ``The explanation of the concept's presence in the image is useful for understanding the prediction.''
\end{enumerate}
We call the question for short, visual pattern, highlighted areas, reasonable presence, and useful explanation.

\subsubsection{Step 3: Final Summary Evaluation}
This concluding section for each image re-presents the original image and model prediction for reference, followed by the comprehensive LLEXICORP summary of all the inputs and concepts generated by ChatGPT. Participants then answer three multiple-choice questions, also using the `Agree', `Not sure', or `Disagree' scale:
\begin{enumerate}
    \item ``The summary contains only information that was already present in the descriptions of the individual concepts.''
    \item ``The summary is helpful for understanding the prediction.''
    \item ``The summary is more helpful than the individual concept descriptions.''
\end{enumerate}

We call the question for short, only existing info, a helpful summary, and more helpful.

The questionnaire concludes with an ``End of Image Evaluation'' section, signaling the completion of the assessment for that particular image.

\section{Discussion}

\begin{table}[tb]
    \centering
    \begin{tabular}{lccc}
    \toprule
     & Agree & Not sure  & Disagree \\
    \midrule
    pattern & 78 & 10 & 11 \\
    highlighted areas & 61 & 13 & 26 \\
    reasonable presence & 67 & 15 & 18 \\
    useful explanation & 65 & 15 & 21 \\
    only existing info & 83 & 10 & 7 \\
    helpful summary & 75 & 15 & 10 \\
    more helpful & 33 & 35 & 33 \\
    \bottomrule
    \end{tabular}
    \caption{Aggregated results (as percentages) for each of the question types.}
    \label{tab:aggr}
\end{table}

\begin{table}[tb]
    \centering
    \begin{tabular}{llccc}
        \toprule
         Question& \makecell{Response\\ concept} & Agree & Not sure & Disagree \\
        \midrule
        \multirow[c]{5}{*}{pattern} & 1 & 94 & 3 & 2 \\
         & 2 & 73 & 10 & 15 \\
         & 3 & 74 & 14 & 11 \\
         & 4 & 77 & 11 & 10 \\
         & 5 & 70 & 12 & 16 \\
        \cmidrule(lr){1-5}
        \multirow[c]{5}{*}{highlights} & 1 & 85 & 9 & 5 \\
         & 2 & 54 & 11 & 34 \\
         & 3 & 58 & 15 & 26 \\
         & 4 & 54 & 13 & 32 \\
         & 5 & 52 & 14 & 33 \\
        \cmidrule(lr){1-5}
        \multirow[c]{5}{*}{presence} & 1 & 92 & 7 & 0 \\
         & 2 & 62 & 16 & 21 \\
         & 3 & 58 & 23 & 18 \\
         & 4 & 63 & 14 & 22 \\
         & 5 & 60 & 14 & 25 \\
        \cmidrule(lr){1-5}
        \multirow[c]{5}{*}{explanation} & 1 & 92 & 5 & 2 \\
         & 2 & 57 & 19 & 23 \\
         & 3 & 56 & 20 & 22 \\
         & 4 & 58 & 15 & 26 \\
         & 5 & 57 & 11 & 30 \\
        \bottomrule
    \end{tabular}
    \caption{Aggregated results (as percentages) for each concept rank separately. The importance of the first concept is much higher than the remaining.}
    \label{tab:byconcept}
\end{table}

\begin{table}[tb]
    \centering
    \begin{tabular}{lccc}
        \toprule
         & Agree & Disagree & Not sure \\
        \midrule
        highlighted areas & 72 & 19 & 10 \\
        reasonable presence & 76 & 12 & 11 \\
        useful explanation & 74 & 14 & 12 \\
        \bottomrule
    \end{tabular}
    \caption{Aggregated results as percentages given that the respondents answered `Agree' to the visual pattern.}
    \label{tab:given_pattern}
\end{table}

\begin{table}[tb]
    \centering
    \begin{tabular}{lccc}
        \toprule
        response & Agree & Disagree & Not sure \\
        \midrule
        reasonable presence & 93 & 2 & 4 \\
        useful explanation & 89 & 4 & 7 \\
        \bottomrule
    \end{tabular}
    \caption{Aggregated results as percentages given that the respondents answered `Agree' to the visual pattern and highlighted areas.}
    \label{tab:given_pattern}
\end{table}

\begin{table}[tb]
    \centering
    \begin{tabular}{lccc}
        \toprule
        response & Agree & Disagree & Not sure \\
        \midrule
        useful explanation & 92 & 3 & 5 \\
        \bottomrule
    \end{tabular}
    \caption{Aggregated results as percentages given that the respondents answered `Agree' to the visual pattern, highlighted areas, and reasonable presence.}
    \label{tab:given_pattern}
\end{table}

Across all evaluated concepts, participants consistently recognized meaningful visual patterns in the representative images (78\% agreement for the \emph{visual pattern} question, only 11\% disagreed, Table~\ref{tab:aggr}).
This confirms that LLEXICORP successfully identifies coherent visual features that participants can understand as recurring elements.
The top-ranked concept, in particular, performed exceptionally well across all metrics (Table~\ref{tab:byconcept}), with over 90\% agreement for both \emph{reasonable presence} and \emph{useful explanation}.
This indicates that the most salient concept provides an interpretable and trustworthy foundation for understanding the model’s prediction.

Another key strength lies in the final summaries (Step~3). Participants strongly agreed (83\%) that the summaries contained only information already present in the individual concept descriptions, confirming that the synthesis step avoided introducing misleading or extraneous explanations.
This number also seems artificially lowered by the fact that some respondents treated the reported relevance number as not included.

Furthermore, 75\% of participants found the summaries helpful for understanding the prediction, suggesting that aggregating concept-level insights into a single narrative generally improves interpretability and provides a more coherent perspective for end users.

The conditional analyses in Tables~\ref{tab:given_pattern}--\ref{tab:given_pattern} reveal a clear hierarchical relationship in explanation acceptance: when participants agreed that a visual pattern was present and correctly localized, agreement on \emph{useful explanation} rose to 92\%. This finding demonstrates that LLEXICORP’s explanations can achieve very high perceived value, provided the underlying saliency and pattern recognition steps are convincing.

\subsection{Areas for Improvement}

Despite these strengths, the evaluation also highlights areas where the current system could be improved. Agreement for \emph{highlighted areas} was comparatively low (61\%), indicating that participants were often uncertain whether the saliency maps correctly indicated where the concept was present in the original image. This uncertainty directly propagated to lower agreement on \emph{reasonable presence} and \emph{useful explanation}, suggesting that clear and accurate localization is a critical bottleneck for explanation trustworthiness. Lower-scoring secondary concepts (ranks 2--5 in Table~\ref{tab:byconcept}) further illustrate this issue: agreement on localization and usefulness drops considerably as concept importance decreases. These less informative concepts risk cluttering the explanation space without significantly improving interpretability.

A second area of potential improvement is the limited incremental value of the final summaries. While they were helpful overall, only 33\% of participants agreed that the summaries were \emph{more helpful} than the individual concept descriptions. This indicates that the summaries largely serve a consolidating function, rather than providing novel interpretive insights.

\section{Conclusion}

In this paper, we introduce LLEXICORP, a modular pipeline for textual explanation of a prediction that follows a long, detailed explainability approach.

We evaluate our approach in a human study, which shows promising results for end-users in building greater trust in AI models.

%\bibliography{bibliography}

\end{document}